\documentclass[runningheads]{llncs}
\usepackage{graphicx}
\usepackage{amsmath}
\usepackage{booktabs}
\usepackage{multirow}

\begin{document}
\title{A Blackbox Model Is All You Need to Breach Privacy: Smart Grid Forecasting Models as a Use Case\thanks{This publication was made possible by NPRP grant 12C-0814-190012 from
the Qatar National Research Fund (a member of Qatar Foundation). The
findings herein reflect the work and are solely the responsibility of the authors.}}

\author{Hussein Aly\inst{1} \and
Abdulaziz Al-Ali\inst{1, 2} \and
Abdullah Al-Ali\inst{1} \and 
Qutaibah Malluhi\inst{1, 2}}
\authorrunning{H. Aly et al.}

\institute{Qatar University, Collage of Engineering, Department of Computer Science and Engineering, Doha, Qatar. 
\email{\{hussein.aly, a.alali, abdulla.alali, qmalluhi\}@qu.edu.qa}\and
KINDI Center for Computing Research\\
}

\maketitle            
\begin{abstract}

This paper investigates the potential privacy risks associated with forecasting models, with specific emphasis on their application in the context of smart grids. While machine learning and deep learning algorithms offer valuable utility, concerns arise regarding their exposure of sensitive information. Previous studies have focused on classification models, overlooking risks associated with forecasting models. Deep learning based forecasting models, such as Long Short Term Memory (LSTM), play a crucial role in several applications including optimizing smart grid systems but also introduce privacy risks. Our study analyzes the ability of forecasting models to leak global properties and privacy threats in smart grid systems. We demonstrate that a black box access to an LSTM model can reveal a significant amount of information equivalent to having access to the data itself (with the difference being as low as 1\% in Area Under the ROC Curve). This highlights the importance of protecting forecasting models at the same level as the data.

\end{abstract}

\section{Introduction}

The use of machine learning and deep learning algorithms has become widespread across various industries due to their ability to analyze large datasets and extract valuable insights. While these models can greatly enhance decision-making processes, there is a concern regarding the potential exposure of sensitive information through their weights \cite{ateniese2015hacking} or even through their predictions under black box settings \cite{zhang2021leakage}. The leaked information can be related to specific records in the dataset (such as membership inference attacks \cite{shokri2017membership}), or to a global property that could reveal private information about the training dataset or the data owner (such as the proportions of specific classes in the training dataset~\cite{zhang2021leakage}). However, previous studies have primarily focused on information leakage from \textit{classification} models, with less emphasis to risks associated with \textit{forecasting} models.

Forecasting models involve creating mathematical models based on historical trends to estimate future values. This technique has found extensive applications ranging from financial market analysis to weather prediction. In the field of power system management, time series forecasting has gained significant traction, particularly in smart grids. Deep learning forecasting models have become essential components of modern power systems. These models (such as Long Short Term Memory) are well-suited for smart grid applications due to their ability to handle time series data \cite{ozcanli2020deep}. The usage of these models enable efficient demand response by forecasting electricity consumption of cities or neighborhoods, facilitating more effective demand response policies \cite{cini2020cluster}. Additionally, forecasting models trained on individual household data can provide feedback to occupants regarding their energy usage patterns, empowering users to adjust their behavior and potentially achieve energy savings. Furthermore, these models contribute to the identification of anomalies in electricity usage, which can indicate electricity theft or faulty appliances \cite{himeur2021artificial}. However, it is important to acknowledge that these models can also introduce significant privacy risks. \par 

The privacy of smart grid users has been a subject of interest over the past decade, as it has impeded the widespread adoption of the smart meter paradigm in certain countries \cite{farokhi2020review}. This is because the fine-grained electricity consumption data collected through smart grids has been shown to reveal sensitive information about household occupants \cite{wang2018deep}. In response to this risk, various techniques have been proposed in the literature to preserve privacy, including data aggregation~\cite{buescher2017two,kserawi2022} and federated learning \cite{rasha2023federated}. However, most of these efforts primarily focus on addressing the privacy risk associated with the electricity data itself, rather than considering the potential privacy implications of models trained on this data. Consequently, this work aims to address these gaps by analyzing the ability of forecasting models to leak global properties in the context of the smart grid. \par

\textbf{Contribution:} This work focuses on analyzing the privacy risks of personalized forecasting models. The goal is to quantify the information an adversary can extract from a black box access to a forecasting model. We propose an attack that allows adversaries to extract global properties about the honest user, such as the number of children, occupants, and the presence of desktop and console devices. We evaluate our attack using the CER dataset collected from the Republic of Ireland \cite{commission2012cer}. The following is a summary of the key contributions of this paper.

\begin{itemize}
    \item To the best of our knowledge, this is the first global properties leakage attack against black box forecasting models.
    \item We assess the effectiveness of the attack in leaking global properties of smart grid electricity data using a real world dataset. 
    \item We demonstrate that the proposed attack can effectively extract multiple global properties under black box settings.
\end{itemize}

\textbf{Organization:} The paper is organized as follows. Section \ref{sec:background} provides basic background information about smart grids and forecasting models. Section \ref{sec:related_work} provides an overview of the related work. Section \ref{sec:meta_class} describes the attack settings and presents a formal modeling of the attack. Then, Section \ref{sec:attack_val} outlines the validation procedure. Section \ref{sec:results} presents and discusses the findings of the attack. Finally Section \ref{sec:conclusion} concludes the paper and discusses some future work directions.

\section{Background}
\label{sec:background}
The smart grid relies on Advanced Metering Infrastructure (AMI), which enables bidirectional communication between the utility provider and the consumer. The AMI has two components: the smart meter, which collects fine-grained electricity consumption data of the user, and a communication network, which enables data transmission between the user and the utility provider in real time. The utility provider can then use this data to provide feedback to the end user to promote more efficient electricity consumption. \par

The adoption of AMI increases the amount of data available to the utility provider. This data can be analyzed by various machine learning and data analysis techniques to extract valuable insights and help optimize the operation and planning of the smart grid by providing accurate forecasting, anomaly detection, fault diagnosis, state estimation, load balancing, demand response and energy management. Data analytics and machine learning can also improve the security and privacy of the smart grid by detecting and preventing cyberattacks, data breaches and unauthorized access.\par

Among all research efforts that use deep learning in smart grid related problems, the Long Short-Term Memory (LSTM) is the most popular model \cite{ozcanli2020deep}. LSTM is a type of recurrent neural network (RNN) that can capture long-term dependencies and temporal patterns in sequential data. This enables LSTM algorithms to provide state-of-the-art performance in many forecasting problems in the smart grid, such as demand prediction and fault detection. Moreover, LSTM models can help in integrating renewable energy with the smart grid, as they can predict the output of solar power or wind power and adjust the supply and demand accordingly. LSTM forecasting models can also support the participation of consumers and prosumers in the smart grid by forecasting their electricity consumption and generation and providing them with optimal pricing and scheduling policies to reduce the overall consumption.

\section{Related work}
\label{sec:related_work}
The privacy concerns associated with smart meters have hindered the widespread adoption of the smart meter paradigm in certain countries \cite{farokhi2020review}. Consequently, extensive research has been conducted to investigate the impact on user privacy resulting from the usage of smart grid technologies. The findings of these studies indicate that the detailed electricity consumption data provided by smart grids can unveil significant user information, including employment status, presence of children, number of occupants, and types of home appliances \cite{montanez2020machine,cui2022realizing,wang2018deep}. Consequently, several approaches have been proposed to mitigate privacy risks associated with smart meter data. These include techniques such as pattern obfuscation using batteries \cite{giaconi2017smart}, privacy-preserving aggregation \cite{buescher2017two,kserawi2022}, and the application of federated learning techniques \cite{rasha2023federated}.  Resulting in a reduction on the privacy threat of the smart meters. \par 

However, the majority of these efforts try to protect the user’s privacy from being exposed by the raw electricity signal. While this is one of the major threat actors to user privacy, it is not the only one. An adversary with access to a model trained on the user data can potentially leak a significant amount of information. There are multiple ways to extract information from machine learning or deep learning models. One of the early attempts was performed by Ateniese et al. \cite{ateniese2015hacking}, where the authors proposed the usage of meta classifiers to extract significant properties from classification models. Significant properties represent a property that is correlated with the training data but not part of the data itself. For example, in a dataset of voice conversations, a significant property can be the accent of the recorded persons in the audio files. They demonstrated that an adversary with a white box access to a classification model and a dataset with the same distribution as the training data can potentially extract multiple information regarding the training data, such as the accent of users in the datasets, or the percentage of male or female users. \par 

The limitation of the Ateniese scheme is that it assumes a white box access to a classification model, which can be difficult to achieve. In a more recent work, Zhang et al. \cite{zhang2021leakage} addressed this limitation. The authors proposed a black box property leakage attack against a classification model. The authors assumed that an adversary would have a black box access to the model, plus knowledge of the training parameters and a dataset sampled from a similar distribution to the target dataset. Their proposed attack is based on a modified version of the Ateniese attack, where the adversary uses posterior probabilities (prediction confidence values) as an attack vector instead of using the model itself. In their attack, the adversary trains multiple shadow models with different distributions of the target properties. For example, assuming that the target property is the percentage of males to females in the dataset, the adversary would train a set of shadow models for each possible distribution of males to females. Then, the adversary would build a meta classifier on the posterior probability vectors returned by the black box shadow model, and their corresponding distribution of the target variable. The meta classifier can then be used to predict the distribution of the target variable from the posterior probability of any black box classification model. This attack is however limited to models that produce posterior probabilities as outputs.\par 

The previous works demonstrated the possibility of extracting significant properties from machine learning models. However, most of the work in this field focused on classification models, and thus may not be applicable to forecasting models. Forecasting is a popular problem in several critical domains such as the smart grid where most approaches use LSTM/RNN algorithms \cite{ozcanli2020deep}. One of possible attacks against LSTM/RNN models are model extraction attacks, in which an adversary is able to train a replica of the target model. Takemura et al. \cite{takemura2020model} proposed a scheme for efficiently conducting model extraction attacks against black box access to LSTM models. The attack uses student-teacher protocol to train an RNN model as a replica of a more sophisticated LSTM model. The attack assumes that the adversary has a black box access to the LSTM with an access to part of the training data. The authors showed that you can efficiently clone a forecasting model using only a black-box access and part of the training data. In contrast to their technique, our proposal focuses on revealing information about the used training data rather than the model. The related work indicate that there is a gap in studying the potential effect on the user privacy when an adversary obtains a black box access to a user's forecasting model. To address this gap, we develop a modified version of the meta classification attack that is tailored for attacking forecasting models. 

\section{Meta Classification Attack}
\label{sec:meta_class}
This section provides an overview of the proposed meta classification attack. The aim of the attack is to leak significant properties about the training data using a black box access to a forecasting model. First, we will discuss the attack settings and assumptions. Then we provide a detailed description of the attack steps. 

\subsection{Attack Setting}
\label{sec:attack_sett}

The goal of the adversary in this attack is to extract significant properties about the honest user's training data. The significant properties are not explicitly part of the training data but can be inferred from it. For example, in the context of a smart grid where the training data is the user's electricity consumption, a significant property can represent the number of appliances in the user's household or the number of house residents.\par

More formally, let us assume an adversary denoted as $E$ whose goal is to leak one or more sensitive properties $S$ of the honest user $H$ using black-box access to a forecasting model $f_h$ trained on the honest user's private data $D_h$. In order for the attack to be effective, the sensitive properties $S$ need to be correlated with the honest user data ($S \approx D_h$). In other words, given $D_h$, one can extract information about $S$. \par 

The adversary is assumed to have access to domain auxiliary data denoted as $D_{aux}$ that are of a similar type to the honest user data. This assumption is applicable in real-life scenarios as such data can be publicly available. For instance, in the context of a smart grid network, $D_{aux}$ can represent electricity consumption data from public datasets. Additionally, we assume that the adversary knows the type of forecasting model $f_h$ used by the honest user (e.g., LSTM, RNN ,etc.) but does not know the training hyperparameters (e.g., number of nodes, learning rate, scaling function, etc.) and does not have access to the model's weights. This assumption is valid, as the general type of the model is not sensitive information and is often publicly available. \par

\subsection{Attack Description}
\label{sec:attack_desc}

The proposed attack consists of two stages: an offline preprocessing stage and an active attack stage. The bulk of the computation cost needed to run the attack happens during the offline preprocessing stage, which reduce the time needed to execute the attack. Next we will describe both the offline and the active stages. 

\begin{figure}[t]
\includegraphics[width=\textwidth]{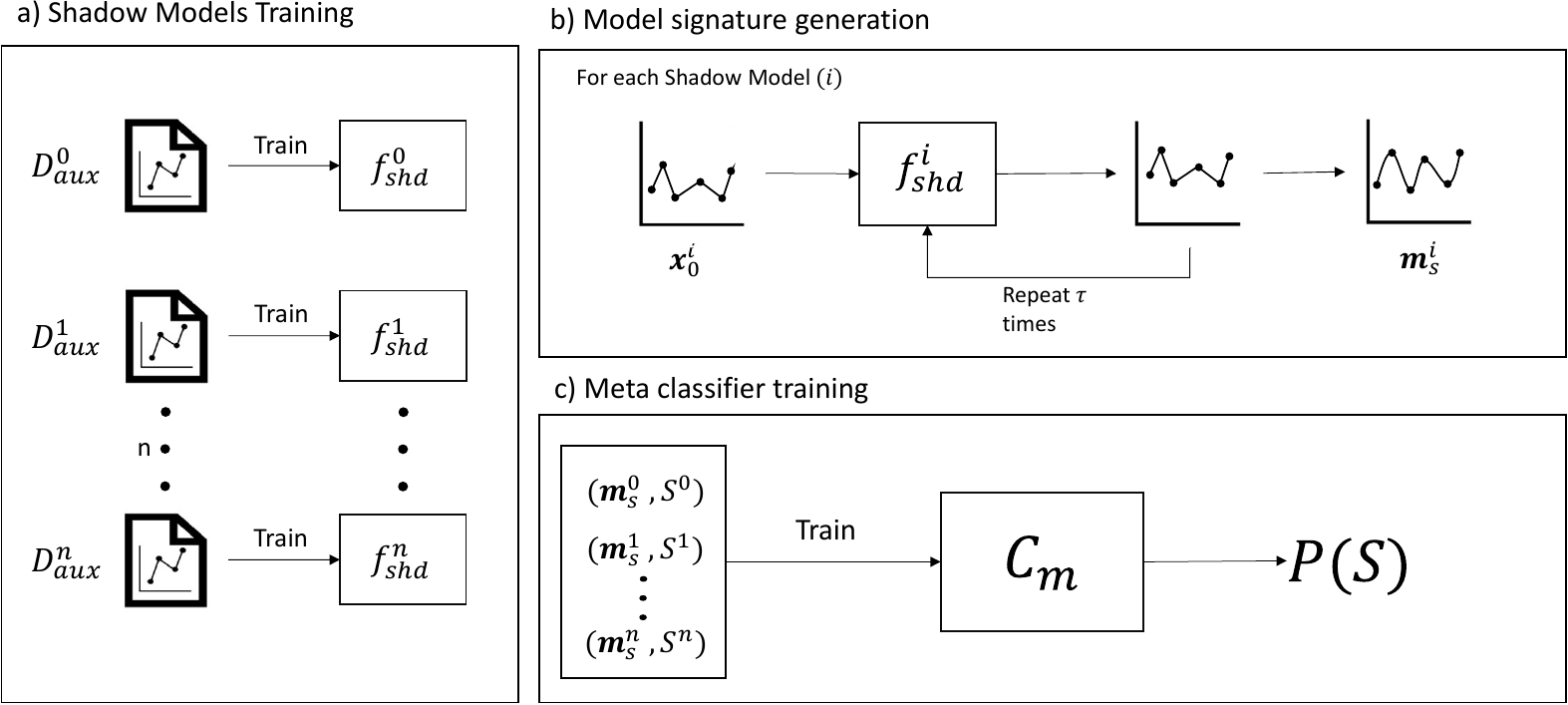}
\caption{Overview of the offline stage: Figure a) illustrates the training of shadow models, while Figure b) depicts the generation of model signatures using the generated shadow models. Figure c) showcases the process of training meta classifiers on the generated model signatures.} \label{fig:offline_stage}
\end{figure}

\textbf{Offline Preprocessing Stage}. This stage consists of three steps. The first step is shadow model training, where an adversary trains multiple shadow forecasting models; one $f_{shd}$ for each subject in $D_{aux}$. Assuming that $D_{aux}$ contains $n$ subjects, the adversary needs to train $n$ shadow models $f_{shd}$. Figure \ref{fig:offline_stage}:a demonstrates an overview of this step. Moreover, since we assume that an adversary does not have access to the training hyperparameters of the honest user model $f_h$ (e.g., number of nodes, learning rate, training epochs, etc.), the adversary may perform hyperparameter tuning on $D_{aux}$ to determine the optimal set of parameters.

The second step is model signature generation, where a model signature $\mathbf{m}_{shd}$ is generated for each shadow model $f_{shd}$ by recursively applying $f_{shd}$ on an initially random input vector $\mathbf{x}_0$ through $\tau$ recursive operations as shown in figure \ref{fig:offline_stage}:b. This operation is represented by the following equations 

\begin{equation}
\label{eq:model_sig_gen}
\begin{aligned}
\mathbf{x}_0 &= U(0, 1) \quad \text{for } |\mathbf{x}_0| = w \\
\mathbf{x}_n &= f(x_{n-1}) \quad \text{for } 0 < n < \tau \\
m_s &= \mathbf{x}_\tau
\end{aligned}
\end{equation}

The model signature generation operation is also used in the online stage, thus in the previous equation we used $f$ instead of $f_{shd}$ to indicate any forecasting model. The last step of the offline stage is meta classifier training, where a meta classifier $C_m$ is trained to learn the relationship between the model signature and the target significant property. Specifically, $C_m$ is trained on pairs $(\mathbf{m}_s, S)$ for each user in $D_{aux}$. This is shown in figure \ref{fig:offline_stage}:c. After the meta classifiers training, the adversary saves the generated meta classifier models in order to use them in the active stage.  

\begin{figure}[t]
\includegraphics[width=\textwidth]{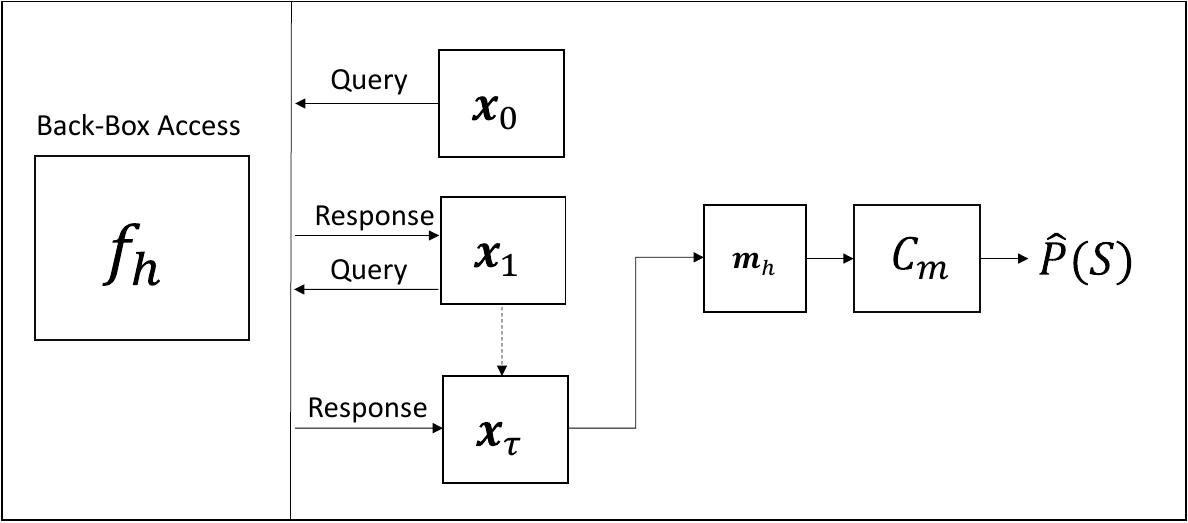}
\caption{Overview of the active attack stage. where $f_h$ represents the honest user forecasting model, $\mathbf{x}_0$ donates a randomly initialized vector, $\mathbf{m}_h$ donates a model signature, and $C_m$ donates the meta classifier, finally \^p(S) represents the probability of the significant property S} \label{fig:online_stage}
\end{figure}

\textbf{Active Attack Stage.} During this stage, the adversary utilizes black box access to the target user model to construct the model signature $\mathbf{m}_h$ for the honest user. This is achieved by applying equation \ref{eq:model_sig_gen}. Then, using the meta classifier $C_m$, the adversary predicts the probability of the significant property for the target user: \^P(S)$ = C_m(\mathbf{m}_h)$. Figure \ref{fig:online_stage} provides an overview of the active attack stage, where the recursive operation needed to generate the model signature is replaced with a query-response relation. Each recursive iteration is modeled as querying the black box model and sending the response back as a new query.

\section{Attack validation}
\label{sec:attack_val}

As a use-case of our attack, we applied it in the context of the smart grid, where the adversary aims to extract information about the honest user by accessing a personalized forecasting model trained on the user's private data. Personalized forecasting models are commonly employed to identify outliers in electricity consumption, which could be indicative of faulty appliances or inefficient usage patterns. Users may utilize systems that offer services for detecting faulty appliances to improve methods for reducing overall energy consumption. In this particular scenario, our objective is to measure the level of information leakage that an adversary can achieve through black box access to such forecasting models. By quantifying this leakage, we can better understand the potential privacy risks.

To evaluate this attack, we utilized the CER Ireland Electricity Customer Behavior Trial dataset \cite{commission2012cer}, which provides half-hourly consumption data for 4,233 households in the Republic of Ireland over an average period of 536 days. We chose this dataset specifically because it includes multiple households, enabling us to effectively evaluate our attack. Additionally, the dataset provides a relatively long duration of data for each household, which assists us in developing personalized models for each household.

From the Irish dataset, we identified 8  properties that we believe an adversary would be interested in knowing from the user data. These properties, listed in Table \ref{tab:cer_prop}, encompass information about the household and its occupants, such as the retirement status of the chief income earner, the number of occupants, and the number of children. Moreover, the identified properties also include information about the appliances in the household, such as the type of cooking facility, and the presence of consoles and desktop computers. Finally, the table includes properties about the house itself, such as the house type and age. One may argue that these properties are not sensitive and can be accessible from looking at the household (such as the house type) or any light weight interaction with the household occupants (such as the number of children's or occupants). This is true assuming that the adversary has physical access to the household. However, if the adversary managed to obtain a remote black box access to the target forecasting model, gaining such information could lead to a potential risk for the user.

\begin{table}[t!]
\centering
\caption{Significant Properties extracted from the CER data set. The counts indicates the number of household for each label. Also,Households with missing information where removed. }
\label{tab:cer_prop}
\resizebox{0.75\textwidth}{!}{%
\begin{tabular}{@{}lll@{}}
\toprule
Question & Label & Count \\ \midrule
Chief income earner retired or not & No & 2947 \\
 & Yes & 1285 \\
Cooking facility type & Not Electrical & 2960 \\
 & Electrical & 1272 \\
Having children & Yes & 1229 \\
 & No & 3003 \\
Living alone & No & 3424 \\
 & Yes & 808 \\
House age & Old (\textgreater{}=30) & 2152 \\
 & New (\textless{}30) & 2077 \\
House type & Detached or bungalow & 2189 \\
 & Semi-detached or terraced & 1964 \\
Number of gaming consoles & None & 2794 \\
 & One or more & 1438 \\
Number of desktop computers & None & 2231 \\
 & One or more & 2001 \\
\end{tabular}%
}
\end{table}

To evaluate the effectiveness of our attack, we establish a comparison with the amount of information that can be extracted directly from the raw electricity data. For this purpose, we employ a 2D ResNet18 model as a baseline model, trained to predict the significant properties listed in table \ref{tab:cer_prop} using the raw consumption data. To ensure a fair and consistent evaluation, we also construct the meta classifier based on the ResNet18 architecture, enabling a direct comparison between our attack and the baseline model. \par 

Furthermore, we sorted the user meter IDs in CER dataset in ascending order and selected the first 80\% of the sorted IDs to comprise the $D_{aux}$ dataset, which is accessible to the adversary. The remaining 20\% of the IDs were used to evaluate the performance of the attack. Although the attack primarily targets a single user (e.g., a single household), we chose to include 20\% of the meters to represent the honest user in order to evaluate the attack's performance across different user types and obtain a more robust understanding of its effectiveness. \par 

For the choice of forecasting model, we employed an LSTM model, which is a popular choice for forecasting in smart grid problems \cite{ozcanli2020deep}. The architecture of the LSTM model consists of two parallel branches: the first branch is fed with electricity consumption data, while the second branch receives time data. Then the output of both branches is concatenated and passed to a final dense layer. This parallel architecture ensures that the model is well aware of the relationship between time and the user's electricity consumption. \par

\subsection{Attack Execution: Offline Preprocessing Stage}

The first step in the offline preprocessing stage is to train the forecasting models. In our attack, we do not assume that the adversary has access to the hyperparameters of the honest user's data. Therefore, for the adversary's model, we performed a hyperparameter search using Bayesian optimization over a validation set of 10 smart meters chosen at random from $D_{aux}$. The adversary's goal during this tuning stage is to find a set of hyperparameters that work best for each household in the validation set. \par 

As for the honest user model, conducting hyperparameter tuning for each individual user is computationally expensive. Instead, we selected a subset of 10 households of the honest user meters and performed tuning for a single set of hyperparameters that provided the best performance on average across the 10 households. Therefore, the same model hyperparameters were used for all honest users. The hyperparameters used include the learning rate, L2 regularization, LSTM layer nodes, fully connected (FC) layer nodes, scaling function (min-max or standard scaling), and the window size of the model input. \par 

After training the LSTM models, we extracted the model signatures based on the procedure described in Section \ref{sec:attack_desc}. For each forecasting model trained by the adversary, we generated $K$ model signatures. As our LSTM models take as input a history of predictions along with the corresponding timestamps, we randomly sampled $K$ dates from the range of available training data. In our experiment, we used $K $$= 100$ to ensure an adequate number of samples that represent the full duration of the training data. Once the model signatures were generated, we trained  ResNet18 models to predict the significant property based on the model signature. This process was repeated for each significant property of interest. The meta-classification models were then saved for use during the active attack stage. \par 

\subsection{Attack Execution: Active Attack Stage}

In this stage, the adversary takes advantage of their access to the forecasting model of the honest user in order to extract information about the significant properties of the user. The execution of this stage follows the approach described in Section \ref{sec:attack_desc}. Initially, the adversary interacts with the honest user's forecasting model and obtains $K$ model signatures by using the $K$ dates values generated in the offline stage. 
Subsequently, these generated model signatures are utilized by the meta classifiers to calculate probabilities for each of the significant properties. It is important to note that regardless of the number of significant properties being leaked,

\subsection{Evaluation Metrics}

The evaluation of this attack is based on the amount of information extracted from the model signatures compared to the baseline model. Therefore, a smaller difference between the attack results and the baseline indicates a better attack. In the best-case scenario, the attack would produce the same performance as the baseline, while in the worst-case scenario, the attack performance would be similar to random guessing. Since the baseline model is a classification problem, the attack is evaluated using the following metrics:

\textbf{Area under the ROC curve (AUC)} : The ROC curve is a graphical plot that measures the trade off between True Positive Rate (TPR) and False Positive Rate (FPR) for a binary classifier on all classification thresholds . The AUC represents the area under this curve, which is used as a measure of the classifier's ability to distinguish between positive and negative instances. A higher AUC indicates a better classifier performance. \par 
\begin{equation}
    TPR = \frac{TP}{TP + FN}
\end{equation}
\begin{equation}
    FPR = \frac{FP}{FP+TN}
\end{equation}
Where $TP$, $FP$, $TN$, $FN$ are the numbers of true positives, false positives, true negatives, and false negatives respectively. 

\textbf{Precision} : indicates the quality of the positive predictions, as it measures correctly predicted positive instances  out of all predicted positive instances. \par 
\begin{equation}
    Precision = \frac{TP}{TP + FP}
\end{equation}

\par 

\textbf{Recall} : indicates the detection rate of all positive samples, as it measures how many positive samples where correctly classified out of all available positive samples. 
\begin{equation}
    Recall = \frac{TP}{TP + FN}
\end{equation}
\par 
\textbf{F1 score} : Is the harmonic mean between precision and recall, which quantifies the balance between precision and recall. 

\begin{equation}
    F1 = 2 \times \frac{(Precision \times Recall)}{(Precision + Recall)}
\end{equation}
\par

For the F1, Precision, and Recall we used the macro averaging in order to give a representative reporting scheme for all class labels. 
By utilizing these metrics, we obtained a comprehensive assessment of the attack's performance in comparison to the baseline model.

\section{Results \& Discussion}
\label{sec:results}
\begin{table}[t]
\centering
\caption{Results of the attack}
\label{tab:attack_results}
\resizebox{0.85\textwidth}{!}{%
\begin{tabular}{cccccc}
\hline
\multirow{2}{*}{Significant Property} & \multirow{2}{*}{Model} & \multirow{2}{*}{AUC} & \multirow{2}{*}{F1} & \multirow{2}{*}{Precision} & \multirow{2}{*}{Recall} \\
 &  &  &  &  &  \\ \hline
\multirow{3}{*}{Chief income earner retired or not} & Baseline & 84.22 & 73.96 & 73.69 & 74.34 \\
 & Random & 50.00 & 49.41 & 50.32 & 50.36 \\
 & Adversary & 74.10 & 65.4 & 65.6 & 67.21 \\ \hline
\multirow{3}{*}{Cooking facility type} & Baseline & 75.66 & 66.8 & 67.95 & 67.03 \\
 & Random & 50.00 & 48.63 & 49.3 & 49.23 \\
 & Adversary & 68.32 & 59.48 & 61.96 & 62.79 \\ \hline
\multirow{3}{*}{Having children} & Baseline & 82.53 & 72.25 & 71.45 & 75.06 \\
 & Random & 50.00 & 46.73 & 50.43 & 50.53 \\
 & Adversary & 76.69 & 68.56 & 68.16 & 69.54 \\ \hline
\multirow{3}{*}{House age} & Baseline & 71.17 & 64.35 & 65.09 & 64.76 \\
 & Random & 50.00 & 50.85 & 50.87 & 50.87 \\
 & Adversary & 63.83 & 60.34 & 60.86 & 60.59 \\ \hline
\multirow{3}{*}{House type} & Baseline & 67.46 & 63.21 & 63.31 & 63.28 \\
 & Random & 50.00 & 49.96 & 50.02 & 50 \\
 & Adversary & 63.68 & 56.76 & 61.37 & 59.31 \\ \hline
\multirow{3}{*}{Living alone} & Baseline & 86.09 & 74.61 & 72.81 & 78.94 \\
 & Random & 50.00 & 46.01 & 49.51 & 49.27 \\
 & Adversary & 80.95 & 69.55 & 69.69 & 71.89 \\ \hline
\multirow{3}{*}{Number of gaming consoles} & Baseline & 80.07 & 70.48 & 71.41 & 70.1 \\
 & Random & 50.00 & 49.58 & 50.33 & 50.37 \\
 & Adversary & 73.96 & 66.26 & 67.44 & 66.05 \\ \hline
\multirow{3}{*}{Number of desktop computers} & Baseline & 69.26 & 63.45 & 63.62 & 63.47 \\
 & Random & 50.00 & 50.47 & 50.52 & 50.51 \\
 & Adversary & 68.22 & 62.27 & 63 & 62.45 \\ \hline
 \hline
 \multirow{3}{*}{Average} & Baseline & 77.06 & 68.64 &  68.67 & 69.62\\
 & Random & 50.00 & 48.95 & 50.16 & 50.14 \\
 & Adversary & 71.44 & 63.58 & 64.76 & 64.98 \\ \hline

\end{tabular}%
}
\end{table}

\begin{figure}[t]
    \centering
    \includegraphics[width=\textwidth]{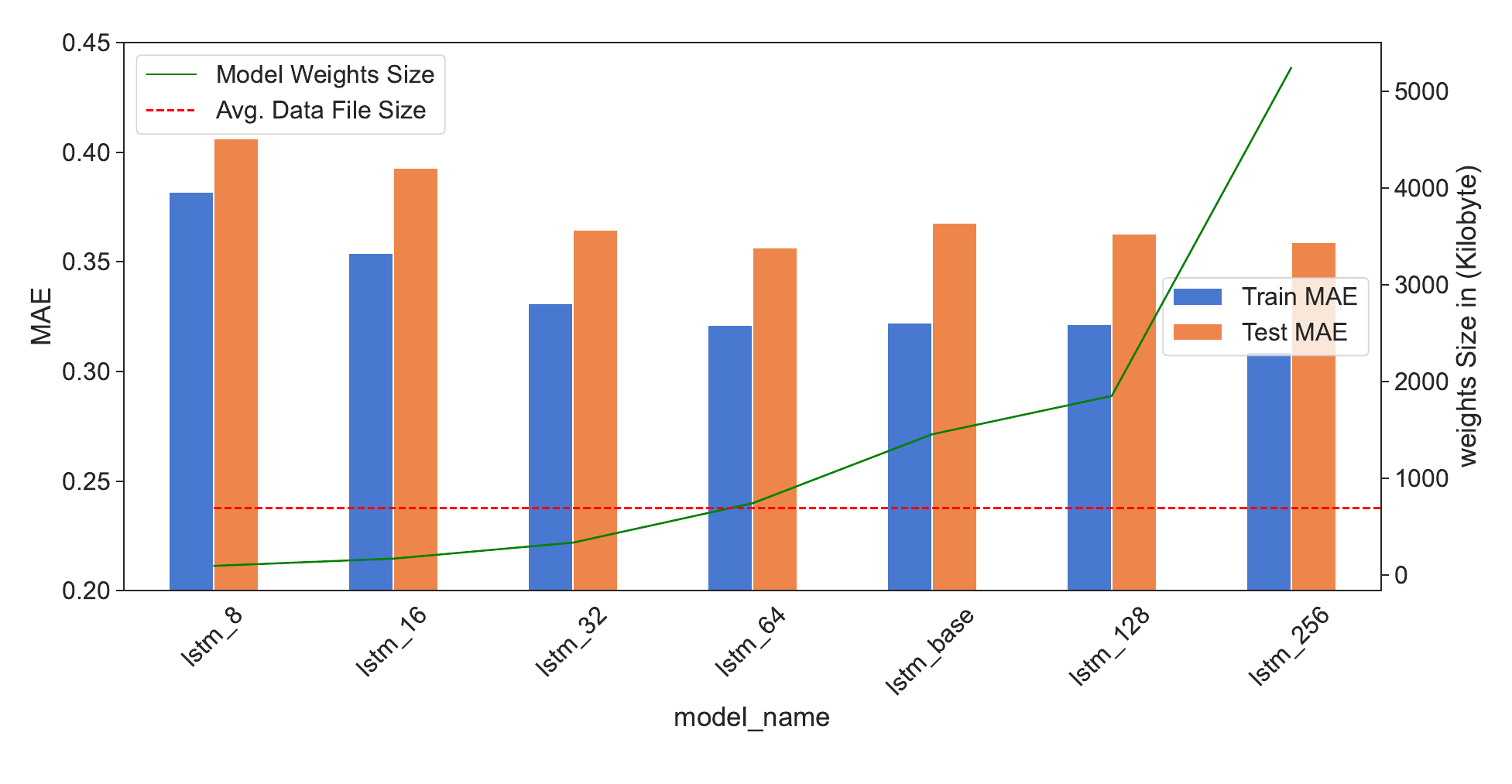}
    \caption{Comparison between the MAE error and model size.}
    \label{fig:size_vs_mae}
\end{figure}

In this work, we aimed to demonstrate that having black box access to a forecasting model can enable an adversary to extract a significant amount of information about the user’s electricity consumption. However, one might argue that if the model size is larger than the size of the user data, then the data may be easier to obtain than the model. Furthermore, the model could easily memorize and leak substantial information, which would make our attack trivial. To address this concern, we carefully selected the LSTM architecture used by the honest user, ensuring that the model size is smaller than the average size of available data per user. We also compared different sizes of LSTM architectures (with different numbers of hidden units) to find the optimal trade-off between model size and performance. \par

Figure \ref{fig:size_vs_mae} depicts the relationship between the model size, forecasting Mean Absolute Error (MAE), and the average data size per household in the CER dataset. The figure illustrates the impact on model performance when varying the number of nodes in the LSTM layer and the fully connected layer from the base model acquired through Bayesian optimization. We denote the base model as LSTM\_base (with 110 LSTM nodes and 174 Fully connected nodes), while other models are denoted as LSTM\_8 to LSTM\_16, where the number of nodes in the fully connected layer is always double the number of nodes in the LSTM layer. For example, In LSTM\_8, the model contains 8 nodes in the LSTM layer and 16 nodes in the Fully connected layer.  \par 

From Figure \ref{fig:size_vs_mae}, we observe that increasing the model size is correlated with a reduction in test MAE. However, the model reaches a point of diminishing returns, where further doubling of the model size results in only minor improvements in accuracy. To ensure that the model used is smaller than the size of the data, we selected LSTM\_32 as the honest user model. \par 

Furthermore, we conducted the attack described in the previous sections on the LSTM\_32 model. The results of this experiment are presented in Table \ref{tab:attack_results}. The table demonstrates that a significant amount of information can be extracted from the black box forecasting model, which closely approximates the amount of information extracted using the private data directly. Despite the honest user deploying a model with a smaller weight size than the data itself, a substantial amount of information can still be leaked. Specifically, the table reveals that the information extracted using the model signatures can closely match the data extracted from the original dataset, with an average difference in F1 score of around 5\%, which can reach as low as 1\% in some properties such as the number of desktops. Also, the table shows that the adversary can obtain a significant advantage over random guessing that could reach up to 30\% increase in the AUC score. \par 

\begin{figure}[t]
    \centering
    \includegraphics[width=\textwidth]{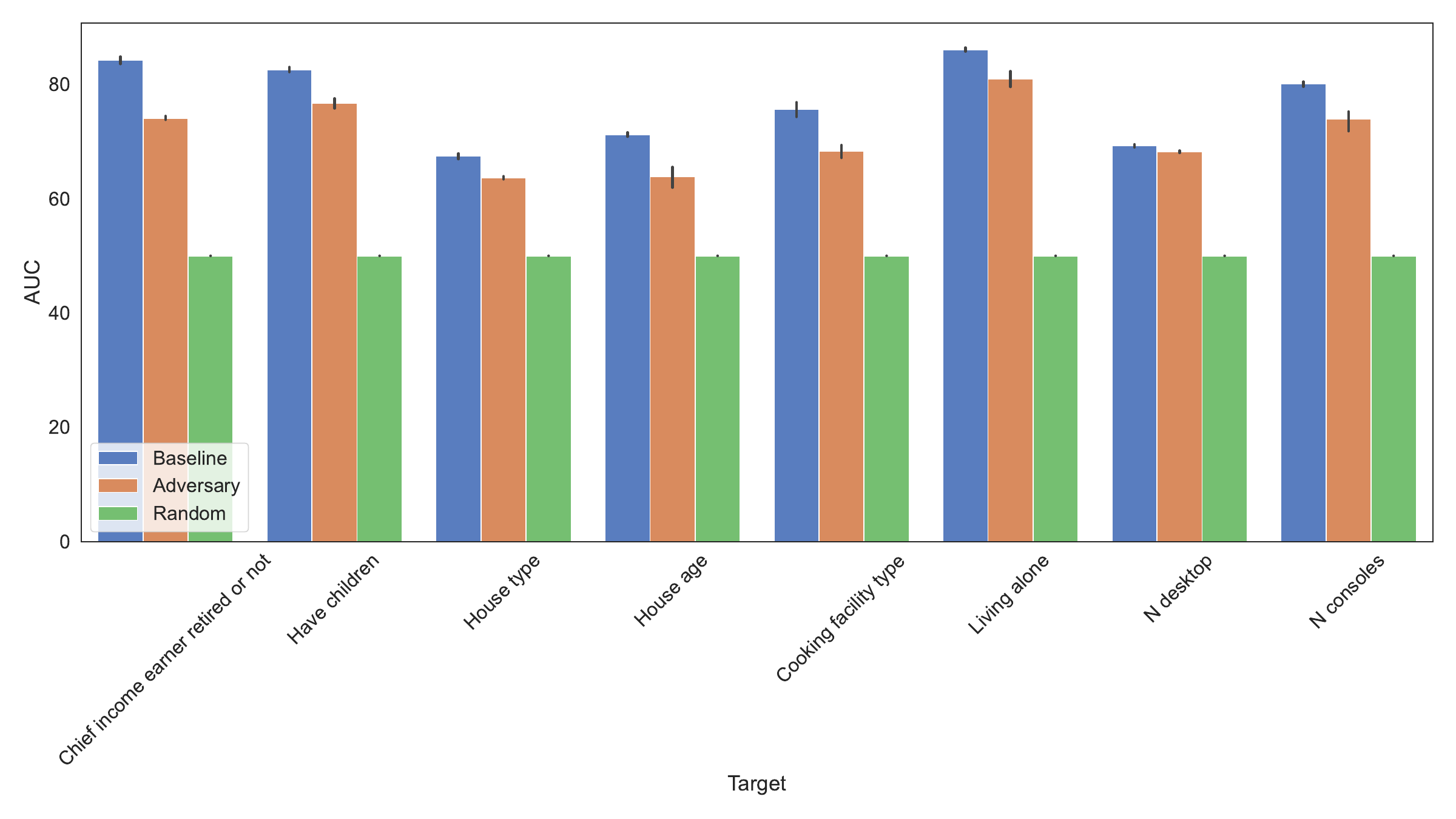}
    \caption{A bar chart representing the AUC scores of the proposed attack compared with the baseline ResNet18 model and random guessing}
    \label{fig:results_auc}
\end{figure}

\begin{figure}[t]
    \centering
    \includegraphics[width=\textwidth]{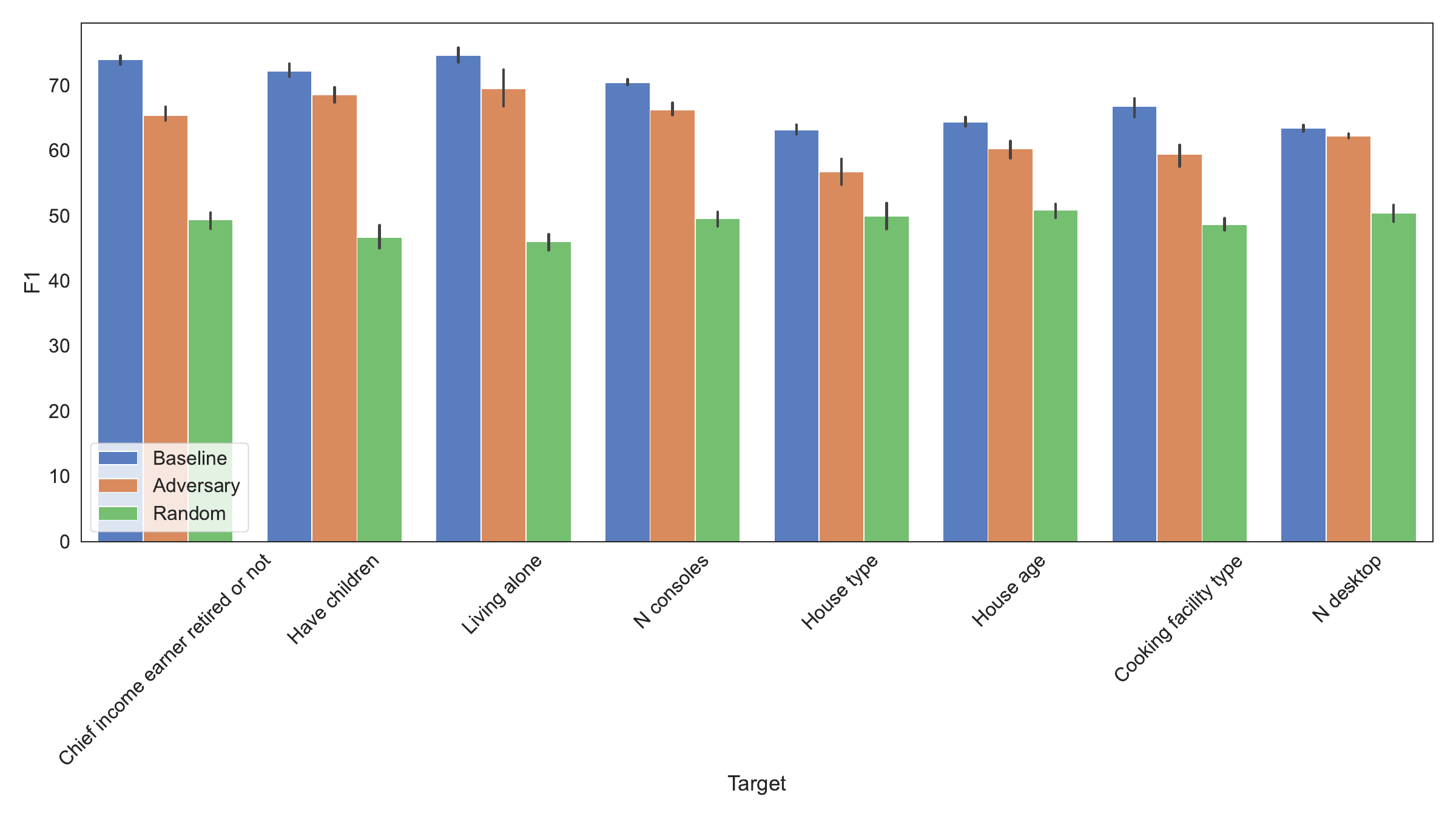}
    \caption{A bar chart representing the F1 scores of the proposed attack compared with the baseline ResNet18 model and random guessing}
    \label{fig:results_f1}
\end{figure}

Moreover, our analysis reveals that the adversary achieves a balance between precision and recall at approximately 65\% for both metrics, which is only 5\% lower than the performance of the baseline model. This indicates that the adversary can successfully extract a significant amount of information from the black box forecasting model, surpassing the performance of a random guessing model by around 15\%. Figures \ref{fig:results_auc} and \ref{fig:results_f1} present bar plots depicting the AUC and F1 score, respectively. These figures clearly illustrate that the adversary consistently maintains a significant advantage over random guessing, while being only a few points away from the baseline performance.

These findings highlight the vulnerability of forecasting models, similar to classification models, to the meta classification attack. Such an attack has the potential to lead to the leakage of sensitive information even when the adversary has only black box access to the model.This paper underscores the importance of implementing robust security measures to protect offline and private forecasting models, as they contain a considerable amount of information comparable to the raw data itself. \par

 \par

\section{Conclusion}
\label{sec:conclusion}
In this paper, we have presented a comprehensive investigation into the vulnerability of black box forecasting models in the context of smart grid applications. Our goal was to assess the potential information leakage that can occur when an adversary gains access to such models. Through our experiments and analysis, we have demonstrated that even with limited knowledge in a black box settings, adversaries can extract a significant amount of information about the user's private data.  \par 

Our research highlights the need for robust security measures in safeguarding offline and private forecasting models. The results of our attack clearly demonstrate that the information extracted from black box forecasting models closely resembles the information obtained from the original private data. This finding underscores the importance of treating forecasting models as potential sources of sensitive information and prompts the development of stronger security protocols to protect user privacy. \par 

Our work contributes to the growing body of research on the security and privacy aspects of machine learning models in critical domains such as the smart grid. By shedding light on the vulnerabilities of black box forecasting models, we aim to raise awareness among researchers, policymakers, and industry professionals about the risks associated with these models and the need for robust defenses. \par 

Moving forward, future research should focus on analyzing potential defense mechanisms such as Differential Privacy and Knowledge Distillation. These techniques can help enhance the security and privacy of forecasting models. Additionally, it is crucial to explore a wider range of attack settings and scenarios to gain a better understanding of the threat landscape. \par

\bibliographystyle{splncs04}
\bibliography{ref}

\begin{thebibliography}{10}
\providecommand{\url}[1]{\texttt{#1}}
\providecommand{\urlprefix}{URL }
\providecommand{\doi}[1]{https://doi.org/#1}

\bibitem{ateniese2015hacking}
Ateniese, G., Mancini, L.V., Spognardi, A., Villani, A., Vitali, D., Felici,
  G.: Hacking smart machines with smarter ones: How to extract meaningful data
  from machine learning classifiers. International Journal of Security and
  Networks  \textbf{10}(3),  137--150 (2015)

\bibitem{buescher2017two}
Buescher, N., Boukoros, S., Bauregger, S., Katzenbeisser, S.: Two is not
  enough: Privacy assessment of aggregation schemes in smart metering. Proc.
  Priv. Enhancing Technol.  \textbf{2017}(4),  198--214 (2017)

\bibitem{cini2020cluster}
Cini, A., Lukovic, S., Alippi, C.: Cluster-based aggregate load forecasting
  with deep neural networks. In: 2020 International Joint Conference on Neural
  Networks (IJCNN). pp.~1--8. IEEE (2020)

\bibitem{cui2022realizing}
Cui, Y., Yan, R., Sharma, R., Saha, T., Horrocks, N.: Realizing multifractality
  of smart meter data for household characteristic prediction. International
  Journal of Electrical Power \& Energy Systems  \textbf{139},  108003 (2022)

\bibitem{commission2012cer}
for Energy Regulation~(CER), C.: Cer smart metering project--electricity
  customer behaviour trial, 2009--2010 (2012)

\bibitem{farokhi2020review}
Farokhi, F.: Review of results on smart-meter privacy by data manipulation,
  demand shaping, and load scheduling. IET Smart Grid  \textbf{3}(5),  605--613
  (2020)

\bibitem{giaconi2017smart}
Giaconi, G., G{\"u}nd{\"u}z, D., Poor, H.V.: Smart meter privacy with renewable
  energy and an energy storage device. IEEE Transactions on Information
  Forensics and Security  \textbf{13}(1),  129--142 (2017)

\bibitem{himeur2021artificial}
Himeur, Y., Ghanem, K., Alsalemi, A., Bensaali, F., Amira, A.: Artificial
  intelligence based anomaly detection of energy consumption in buildings: A
  review, current trends and new perspectives. Applied Energy  \textbf{287},
  116601 (2021)

\bibitem{kserawi2022}
Kserawi, F., Al-Marri, S., Malluhi, Q.: Privacy-preserving fog aggregation of
  smart grid data using dynamic differentially-private data perturbation. IEEE
  Access  \textbf{10},  43159--43174 (2022). \doi{10.1109/ACCESS.2022.3167015}

\bibitem{montanez2020machine}
Monta{\~n}ez, C.A.C., Hurst, W.: A machine learning approach for detecting
  unemployment using the smart metering infrastructure. IEEE Access
  \textbf{8},  22525--22536 (2020)

\bibitem{ozcanli2020deep}
Ozcanli, A.K., Yaprakdal, F., Baysal, M.: Deep learning methods and
  applications for electrical power systems: A comprehensive review.
  International Journal of Energy Research  \textbf{44}(9),  7136--7157 (2020)

\bibitem{rasha2023federated}
Rasha, A.H., Li, T., Huang, W., Gu, J., Li, C.: Federated learning in smart
  cities: Privacy and security survey. Information Sciences  (2023)

\bibitem{shokri2017membership}
Shokri, R., Stronati, M., Song, C., Shmatikov, V.: Membership inference attacks
  against machine learning models. In: 2017 IEEE symposium on security and
  privacy (SP). pp. 3--18. IEEE (2017)

\bibitem{takemura2020model}
Takemura, T., Yanai, N., Fujiwara, T.: Model extraction attacks on recurrent
  neural networks. Journal of Information Processing  \textbf{28},  1010--1024
  (2020)

\bibitem{wang2018deep}
Wang, Y., Chen, Q., Gan, D., Yang, J., Kirschen, D.S., Kang, C.: Deep
  learning-based socio-demographic information identification from smart meter
  data. IEEE Transactions on Smart Grid  \textbf{10}(3),  2593--2602 (2018)

\bibitem{zhang2021leakage}
Zhang, W., Tople, S., Ohrimenko, O.: Leakage of dataset properties in
  multi-party machine learning. In: USENIX Security Symposium. pp. 2687--2704
  (2021)

\end{thebibliography}

\end{document}